\documentclass{article}
\usepackage{spconf,amsmath,amssymb,graphicx, booktabs, microtype}
\usepackage{url}
\usepackage[colorlinks=true, citecolor=green]{hyperref}
\usepackage[capitalize]{cleveref}
\usepackage{makecell}
\usepackage{enumitem}

\title{Event-based Batting Impact Estimation}
\name{
  \begin{tabular}{c}
    Ryotaro Ishida$^{\dagger}$, Wataru Ikeda$^{\dagger}$, Ryosei Hara$^{\dagger}$ \\
    Akemi Kobayashi$^{\ddagger}$, Toshitaka Kimura$^{\ddagger}$, Mariko Isogawa$^{\dagger}$
  \end{tabular}
}
\address{$^{\dagger}$Keio University, $^{\ddagger}$NTT Communication Science Laboratories}

\begin{document}
%
\maketitle
%

\begin{abstract}
Estimating the precise timing of batting impact is crucial for understanding the rapid sensorimotor control.
However, this task is challenging for RGB cameras due to insufficient temporal resolution and motion blur. Similarly, Inertial Measurement Units (IMUs) are impractical for actual matches due to sensor intrusiveness and their limited temporal precision. To overcome these limitations, we propose a novel framework leveraging event-based cameras, which offer microsecond resolution and high dynamic range, to estimate impact timing based on the weighted centroid distance between the detected ball and bat. To address the domain gap between event frames and RGB images that degrades segmentation accuracy, we generate high-density event frames. We then introduce a mask refinement network that leverages these frames and bidirectional mask information, optimized using a novel loss function. Experiments on real-world datasets demonstrate that our method achieves superior accuracy under challenging conditions, including low-light environments and severe occlusions, outperforming baselines by reducing the Mean Absolute Error by approximately 63\%.
\end{abstract}
\vspace{-1mm}

\vspace{-3mm}
\section{Introduction}
\label{sec:intro}
\vspace{-2mm}

In recent years, computer vision has rapidly advanced sports analytics, providing quantitative insights that were previously inaccessible~\cite{Paresh_Ball_tracking}. In sports motion analysis, precise estimation of impact timing is critical not only for decision making~\cite{kobayashi_dicision} but also for evaluating player performance in fast-paced sports~\cite{Hong_E2ESpot,Xarles_TDEED,Caprioli_TennisTiming}. 
Particularly in baseball, accurate temporal localization enables a detailed breakdown of the kinematic chain during the swing phase.
However, the high-speed nature of the game makes this challenging: a ball traveling at 100~km/h covers $\sim$2.8~cm per frame even at 1,000~fps. This spatial deviation is too large to pinpoint the exact moment of contact. Consequently, analysis at even higher temporal resolutions is required to capture sub-millisecond contact durations accurately.

\begin{figure}[t]
    \centering
    \includegraphics[width=\linewidth]{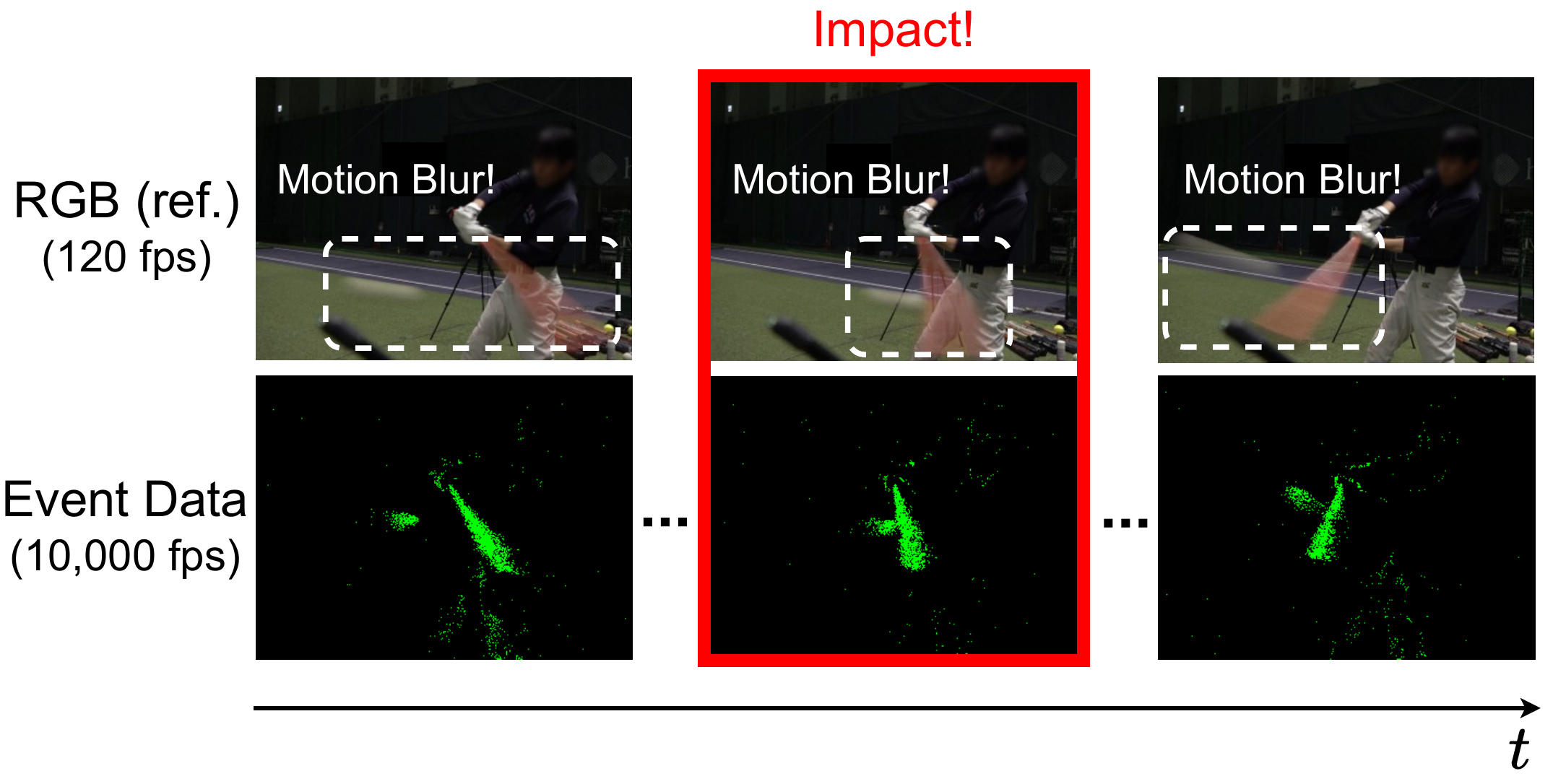}
    \vspace{-10mm}
    \caption{Comparison of sensing modalities.}
    \label{fig:teaser}
    \vspace{-6mm}
\end{figure}

Despite its importance, existing sensing modalities have limitations.
Standard RGB cameras (typically on the order of $10^1$--$10^2$~fps) suffer from insufficient temporal resolution~\cite{Hong_E2ESpot, Xarles_TDEED}.  Fast-moving impacts often occur between frames, and the resulting motion blur obscures the precise boundaries of the ball and bat, making precise timing estimation infeasible from intensity image frames alone. While high-speed cameras can mitigate this, they incur prohibitive hardware costs and generate massive data volumes. Furthermore, unconstrained environments, such as variable lighting in night games, introduce severe challenges for these intensity-based approaches, complicating robust tracking.

Alternative modalities also face fundamental limitations. Acoustic-based methods~\cite{Owens_IndicatedSound, Caprioli_TennisTiming} are susceptible to environmental noise and synchronization latency.
Wearable sensors, such as IMUs, have been employed to detect impact timing via acceleration spikes~\cite{Punchihewa_IMU}.  However, these sensors measure the vibrational response triggered by the collision, rather than directly observing the precise moment of contact initiation. Furthermore, the operational burden of attaching devices to the body or equipment of the athlete makes them intrusive. 
Consequently, there is a strong demand for a robust, non-intrusive vision system capable of sub-millisecond impact estimation under challenging illumination and occlusion.

To overcome these limitations, we leverage event-based cameras~\cite{Gallego_EventRepresentation}. As bio-inspired sensors that asynchronously output pixel-level brightness changes, they offer microsecond-level temporal resolution, high dynamic range, and inherent resistance to motion blur. As shown in Figure~\ref{fig:teaser}, this enables capturing sharp bat and ball motion even in high-speed and low-light scenarios.
However, utilizing event data is challenging due to its spatial sparsity and unique noise. Although recent foundation models trained on massive RGB datasets excel in video domains, they are inherently designed for dense RGB textures. This mismatch results in performance degradation when applying them directly to event frames. Existing event-based detection and segmentation methods~\cite{Gehrig_RVT, Chen_EventSAM} face critical constraints. They typically rely on events accumulated over long time windows (e.g., 10--60~ms). Consequently, these methods fail to produce reliable masks during the sparse, sub-millisecond impact interval. Furthermore, approaches like EventSAM~\cite{Chen_EventSAM} lack explicit temporal modeling, leading to unstable mask predictions that deviate significantly from previous frames.

To address this, we propose the first event-based impact estimation framework capable of operating at 10,000~fps. Our approach follows a coarse-to-fine strategy that effectively adapts robust RGB segmentation models to the high-speed event domain. First, to bridge the domain gap, we specially construct dense event frames even at this high temporal resolution. Since independent frame-wise predictions lack temporal consistency, we employ a bidirectional tracking mechanism to propagate stable masks across time. Subsequently, we introduce a mask refinement network. This network uses the coarse mask as a prior and leverages high-frequency event edges to sharpen object boundaries at the pixel level, maintaining precision even under occlusions.
Finally, to estimate the impact timing, we identify the frame where the distance between the weighted centroids of the ball and bat is minimized.
We validate our method on a real-world dataset including challenging low-light and occluded scenarios.

The main contributions of this paper are summarized as follows:
\textbf{1)} We are the first to propose the task of estimating the impact timing between a fast-moving ball and a returning bat at high temporal resolution using event-based sensing.
\textbf{2)} To this end, we propose a framework that estimates impact timing based on the weighted centroids of the ball and bat. We also introduce a coarse segmentation module that temporally estimates the mask regions of the ball and bat in a bidirectional manner, as well as a mask refinement network that further refines these mask regions for more accurate estimation.
\textbf{3)} As this is a novel task, we also construct the first event-based dataset dedicated to high-speed baseball analysis and demonstrate the robustness of our method in challenging real-world environments, including low-light conditions and severe occlusions.

\begin{figure*}[t]
  \centering
  \includegraphics[width=\linewidth]{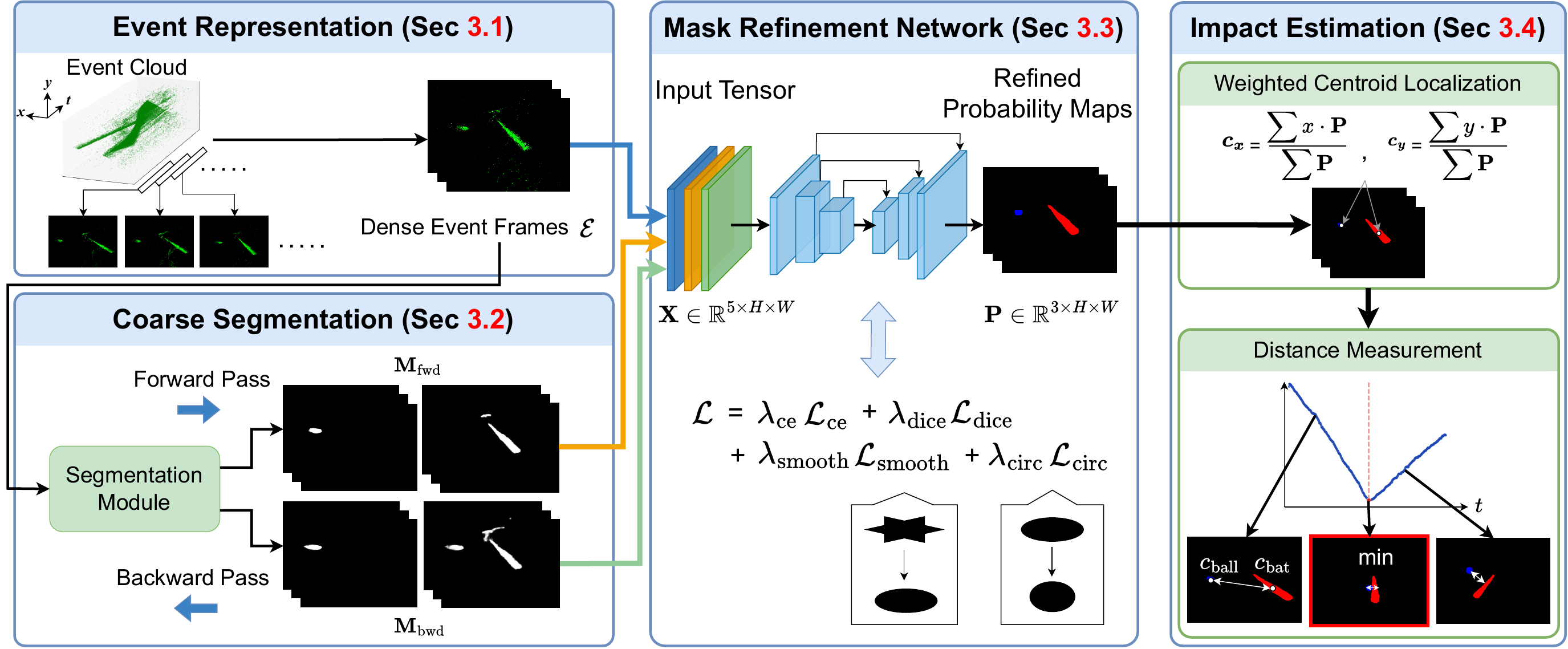}
  \vspace{-6mm}
  \caption{The overview of proposed method}
  \label{fig:overview}
  \vspace{-4mm}
\end{figure*}

\vspace{-4mm}
\section{Related Work}
\label{sec:related_work}
\vspace{-2mm}

\subsection{Impact Timing Estimation for Sports}
\label{ssec:impact_timing_estimation_for_sports}
\vspace{-2mm}

Precise impact timing estimation is crucial for tasks ranging from video-to-audio generation~\cite{Owens_IndicatedSound} to sports performance analysis ~\cite{Hong_E2ESpot,Xarles_TDEED,Caprioli_TennisTiming}. While RGB-based ``Action Spotting" methods~\cite{Hong_E2ESpot,Xarles_TDEED} have advanced, they remain fundamentally constrained by standard frame rates. Specifically, the exact moment of impact is either entirely missed between frames or obscured by severe motion blur. Consequently, existing approaches must often rely on auxiliary modalities like IMUs to estimate impact timing ~\cite{Punchihewa_IMU}. 

In contrast, event-based cameras offer microsecond-level temporal resolution, showing great success in high-speed motion analysis such as video interpolation~\cite{Tulyakov_TimeLens, Kim_EventInterpolation} and robust object detection~\cite{Gehrig_RVT}. Most recently, Kase \textit{et al}.~\cite{Kase_TennisImpact} pioneered the use of event cameras to locate impact positions in tennis. However, their timing estimation relies on simple heuristics based on event counts and polarity changes. Such raw signal metrics are highly susceptible to environmental noise and occlusion. To overcome these limitations, this paper leverages a segmentation-based approach coupled with bidirectional tracking. By enforcing temporal consistency from both directions, our method ensures robust detection even under such challenging conditions.

\vspace{-4mm}
\subsection{Object Boundary Extraction for Contact Detection}
\label{ssec:object_boundary_extraction_for_contact_detection}
\vspace{-2mm}
Accurate impact timing estimation requires precise boundary localization, as slight mask deviations cause significant errors during trajectory intersection. Thus, robust boundary extraction is a prerequisite for this task.
Foundation models for segmentation, such as SAM~3~\cite{SAM3}, have demonstrated remarkable zero-shot generalization capabilities for unseen objects and complex scenes. Various refinement approaches~\cite{Cheng_CascadePSP, Ke_HQSAM} further aim to enhance mask boundary precision. However, these methods predominantly rely on standard camera intensity frames. In high-speed impacts, motion blur obscures the object edges, making contact detection difficult with RGB data alone.

Event-driven adaptations like EventSAM~\cite{Chen_EventSAM} incorporate high temporal resolution into the segmentation pipeline. However, since these methods rely on accumulating events to form structural features, they face a critical trade-off in high-speed impact scenarios. Specifically, extending the integration window induces motion blur, whereas narrowing it results in spatially sparse inputs that cause segmentation breakdown. Furthermore, these methods typically lack mechanisms for temporal consistency, making them unstable for tracking fast-moving objects.
To resolve these limitations, we propose a framework that leverages the semantic robustness of large-scale RGB segmentation models. To bridge the domain gap, we convert raw event streams into dense event frames that maintain sufficient structural density even at high temporal resolutions (10,000 fps). This approach allows us to utilize the RGB model's generalization capability for robust tracking, effectively decoupling feature density from the integration time constraints.

\vspace{-4mm}
\subsection{Vision Datasets for Sports}
\label{ssec:datasets}
\vspace{-2mm}

While event-based vision shows promise in sports, such as the table tennis dataset by Alberico \textit{et al}.~\cite{Alberico_PingPong}, existing benchmarks focus on trajectory tracking rather than microsecond-level impact mechanics. Conversely, traditional baseball datasets like MLB-YouTube~\cite{Piergiovanni_MLBDataset} rely on RGB video, which suffers from temporal aliasing unsuitable for precise contact analysis. Therefore, we introduce the first public dataset combining event-based vision with high-speed baseball impacts, capturing critical dynamics across diverse environments.

\vspace{-4mm}
\section{Proposed method}
\label{sec:proposed_method}
\vspace{-2mm}

We propose a coarse-to-fine impact timing estimation framework that integrates the high temporal resolution and high dynamic range of an event-based camera with SAM~3.  The proposed method consists of four main stages: Event Representation, Coarse Segmentation, Mask Refinement, and Impact Estimation.  The Event Representation stage converts the raw event stream into a dense event frame sequence $\mathcal{E}$. The Coarse Segmentation stage then employs SAM~3 to obtain initial coarse masks $\mathbf{M}_{\text{fwd}}$ and $\mathbf{M}_{\text{bwd}}$.  The Mask Refinement stage utilizes a U-Net-based network, taking the event frames $E_k$ and coarse masks $\mathbf{M}_{\text{fwd}}$, $\mathbf{M}_{\text{bwd}}$ as inputs to generate refined probability maps $\mathbf{P}_{\text{ball}}$, $\mathbf{P}_{\text{bat}}$, and $\mathbf{P}_{\text{background}}$.  The Impact Estimation stage determines the contact moment between the ball and the bat based on the spatiotemporal analysis of these refined probability maps.

\vspace{-4mm}
\subsection{Event Representation}
\label{ssec:event_representation}
\vspace{-2mm}

Let $\mathcal{S} = \{e_i\}_{i=1}^{N}$ be the asynchronous event stream (often referred to as an event cloud), where each event $e_i = (x_i, y_i, t_i, p_i)$ encodes the pixel location, timestamp, and polarity~\cite{Gallego_EventRepresentation}.
To capture the impact timing with high precision, we convert $\mathcal{S}$ into a dense frame sequence $\mathcal{E} = \{E_k\}_{k=1}^{K}$, where each frame $E_k \in \mathbb{R}^{1 \times H \times W}$.
We set the frame interval to $\Delta t = 0.1$~ms, corresponding to a temporal resolution of 10,000~fps.
To ensure sufficient pixel density, we apply a sliding window accumulation. Specifically, the $k$-th frame $E_k$ aggregates positive polarity events ($p_i=1$) within the time window $[t_k - T_{\text{win}}, t_k)$ to prevent trailing streaks from degrading contour precision, where $t_k = k \Delta t$ and $T_{\text{win}} = 1.0$~ms is the accumulation duration.
This approach yields dense event frames with enhanced ball and bat contours, providing sufficient structural information for segmentation while maintaining fine-grained temporal resolution.

\vspace{-4mm}
\subsection{Coarse Segmentation}
\label{ssec:coarse_segmentation}
\vspace{-2mm}

To obtain initial segmentation masks, we prompt SAM~3 with ball/bat point coordinates.
Since the model expects standard video rates, we input the 10,000~Hz event frames $\mathcal{E}$ as a time-stretched 30~fps sequence, aligning apparent motion with the pre-trained priors of SAM~3.
To mitigate occlusions and mask merging at impact, we employ a bi-directional inference strategy.
Processing $\mathcal{E}$ in both temporal directions yields forward and backward probability maps $\mathbf{M}_{\text{fwd}}, \mathbf{M}_{\text{bwd}} \in \mathbb{R}^{2 \times H \times W}$ for the ball and bat.

\vspace{-4mm}
\subsection{Mask Refinement Network}
\label{ssec:mask_refinement_network}
\vspace{-2mm}

\textbf{Network Architecture.} While the initial masks generated by SAM 3 are robust, they remain general-purpose and are not specifically optimized for the challenges arising from object collisions, such as the merging of masks or identity switching during the contact between the ball and the bat.  To address this, we introduce a Refinement Network designed to sharpen object boundaries by leveraging the initial masks as spatial priors.  We employ a U-Net~\cite{Ronneberger_unet} architecture with a ResNet-18~\cite{He_resnet} backbone as the encoder. 

The input to the network is defined as a tensor $\mathbf{X} \in \mathbb{R}^{5 \times H \times W}$, constructed by concatenating the event frame $E_k$ with the forward and backward probability maps $\mathbf{M}_{\text{fwd}}$ and $\mathbf{M}_{\text{bwd}}$.
This configuration enables the network to simultaneously exploit high-frequency edge cues from the event stream and bidirectional spatiotemporal priors provided by the initial tracking.
Finally, the network outputs refined probability maps $\mathbf{P} \in \mathbb{R}^{3 \times H \times W}$, representing the pixel-wise likelihoods for the ball, bat, and background.\\
\textbf{Loss Function.} To ensure pixel-level classification accuracy while guaranteeing the geometric consistency of the object masks, we employ a composite loss function $\mathcal{L}$, defined as:
\begin{equation}
    \mathcal{L} = \lambda_{\text{ce}} \mathcal{L}_{\text{ce}} + \lambda_{\text{dice}} \mathcal{L}_{\text{dice}} + \lambda_{\text{smooth}} \mathcal{L}_{\text{smooth}} + \lambda_{\text{circ}} \mathcal{L}_{\text{circ}},
    \label{eq:total_loss}
\end{equation}
where $\lambda$ terms denote the balancing weights. $\mathcal{L}_{\text{ce}}$ represents the Weighted Cross-Entropy Loss and $\mathcal{L}_{\text{dice}}$ is the Dice Loss~\cite{dice_loss}. These standard terms maximize fundamental segmentation performance. To address the significant class imbalance between the background and the small ball object, we assign higher weights to the ball class in $\mathcal{L}_{\text{ce}}$.

To mitigate mask fragmentation caused by event noise, we introduce a Smoothness Loss ($\mathcal{L}_{\text{smooth}}$). This term acts as a regularizer to suppress isolated noise pixels by minimizing the anisotropic total variation of predicted probability maps $\mathbf{P}$. It is computed as the mean absolute difference between adjacent pixels:

\begin{equation}
    \mathcal{L}_{\text{smooth}} = \frac{1}{N} \sum_{x, y} \left( |\mathbf{P}_{x+1, y} - \mathbf{P}_{x, y}| + |\mathbf{P}_{x, y+1} - \mathbf{P}_{x, y}| \right),
\end{equation}

where $N$ is the total number of pixels.

Furthermore, we impose a shape constraint specifically on the ball class using a Circularity Loss ($\mathcal{L}_{\text{circ}}$). Given that the projection of the ball is circular, we utilize the isoperimetric ratio to enforce compactness. We approximate the perimeter $C$ using the gradients of the probability map $\mathbf{P}_{\text{ball}}$ and the area $A$ as the sum of probabilities. The loss is defined as:
\begin{equation}
    \mathcal{L}_{\text{circ}} = \frac{1}{N}\frac{C^2}{4\pi A + \epsilon}, \, \text{where } C = \sum |\nabla \mathbf{P}_{\text{ball}}|, \ A = \sum \mathbf{P}_{\text{ball}}.
\end{equation}
Here, $\epsilon$ is a small constant introduced for numerical stability.
Minimizing this ratio encourages the predicted mask to approach a perfect circle, for which the ratio is 1, thereby preventing physically implausible deformations.

\vspace{-2mm}
\subsection{Impact Estimation}
\label{ssec:impact_estimation}
\vspace{-2mm}

\textbf{Centroid Localization and Distance Measurement.} Based on the refined probability maps, we estimate object positions using weighted centroids. This approach avoids quantization errors inherent in binary mask centers. The centroid $\mathbf{c} = (c_x, c_y)$ for each object is computed as:
\begin{equation}
    c_x = \frac{\sum_{i,j} x \cdot \mathbf{P}(i,j)}{\sum_{i,j} \mathbf{P}(i,j)}, \quad c_y = \frac{\sum_{i,j} y \cdot \mathbf{P}(i,j)}{\sum_{i,j} \mathbf{P}(i,j)}
\end{equation}
The relative distance $d(t)$ is defined as the Euclidean distance between centroids $\mathbf{c}_{\text{ball}}(t)$ and $\mathbf{c}_{\text{bat}}(t)$. This strategy combines temporal consistency with sub-pixel precision, yielding a coherent trajectory sequence.\\
\textbf{Timing Determination.}
The impact corresponds to the instant of maximum proximity between the two objects. Since the rotational mechanics of a bat swing induce non-linear distance changes, linear fitting approaches like RANSAC are often insufficient. Therefore, leveraging the 10,000~Hz resolution, we determine the impact timing $t_{\text{impact}}$ by directly minimizing the relative distance:
{
\setlength{\abovedisplayskip}{4pt}
\setlength{\belowdisplayskip}{4pt}
\begin{equation}
    t_{\text{impact}} = \operatorname*{arg\,min}_{t} d(t)
\end{equation}
}
This model-free approach effectively utilizes the high-frequency sampling to capture the precise collision frame without relying on geometric assumptions that may not hold during complex swing dynamics.

\vspace{-2mm}
\section{Experiment}
\label{sec:experiment}
\vspace{-2mm}

\subsection{Experimental Setup and Dataset}
\label{ssec:experimental_setup_and_dataset}
\vspace{-2mm}

\textbf{Data Collection.}
Since no public event-based dataset exists for high-speed baseball analysis, we constructed a novel dataset utilizing a DAVIS346~\cite{davis346} event-based camera ($346 \times 260$).
A pitching machine was positioned at a standard distance of about $18.44$~m from the home plate, pitching to three batters: a right-handed former professional player and two left-handed participants.
The impact sequence of the batter was recorded from the side view.
The camera position was slightly adjusted per session to optimize the field of view.\\
\textbf{Dataset.}
Table~\ref{tab:dataset} details the seven scenarios collected to evaluate robustness.
The dataset covers varying conditions: illumination (low: $\approx 18$~lux, mod: $\approx 435$~lux, high: $\approx 1491$~lux), occlusion (safety net), ball material (soft/hard), pitch type (fast/slider), and speed ($100$--$145$~km/h).
In total, the dataset comprises 40 clips and 23,779 event frames.\\
\textbf{Ground Truth Annotation.}
Initially, we attached an IMU to the bat knob, sampling at 1000 Hz, synchronized via an LED trigger. However, preliminary analysis revealed a variable latency of 0.4--7.1 ms in the IMU signal, rendering it insufficient for validating our sub-millisecond estimation. Therefore, we established the Ground Truth through manual inspection of the event stream. To minimize subjective bias and ensure reliability, three human annotators independently pinpointed the exact timing of contact. The final ground truth was defined as the average of these three annotations.\\
\textbf{Evaluation Protocol.}
We adopted a two-stage strategy to evaluate the generalization capability of our method.
First, for the low-light datasets (Scenarios 1--3), we employed a Leave-One-Scenario-Out (LOSO) cross-validation.
In each fold, one scenario was used for testing, while the remaining two were combined for training.
Second, to assess robustness against unseen pitch speeds and types, Scenarios 4--7 were used exclusively for testing.
For these scenarios, the inference was performed using the model trained on the union of Scenarios 1--3.
In all phases, the dataset set was split into training and validation subsets with an 8:2 ratio to monitor convergence.

\begin{table}[t]
\caption{Detailed configuration and statistics of our dataset.}
\label{tab:dataset}
\centering
\resizebox{\columnwidth}{!}{
\begin{tabular}{c|ccccccccc}
\Xhline{0.8pt}
\textbf{Scen.} & \textbf{Lux} & \textbf{Net} & \textbf{Ball} & \textbf{Type} & \textbf{Speed} & \textbf{\# Clips} & \textbf{\# Frames} \\
\hline
1 & Low & No & Soft & Fast & 100 & 5 & 2,080 \\ 
2 & Low & Yes & Soft & Fast & 100 & 4 & 1,440 \\
3 & Low & Yes & Hard & Fast & 100 & 6 & 2,270 \\
4 & Mod & Yes & Soft & Fast & 145 & 9 & 7,290 \\ 
5 & Mod & Yes & Soft & Slider & 130 & 6 & 4,510 \\ 
6 & High & Yes & Soft & Fast & 145 & 5 & 3,235 \\
7 & High & Yes & Soft & Slider & 130 & 5 & 2,954 \\
\hline
\multicolumn{6}{r|}{\textbf{Total}} & \textbf{40} & \textbf{23,779} \\
\Xhline{0.8pt}
\end{tabular}
}
\vspace{-4mm}
\end{table}

\vspace{-2mm}

\subsection{Baseline}
\label{ssec:baseline}
\vspace{-2mm}

To rigorously evaluate the contributions of each component in our framework, we compare our method against a baseline. For a fair comparison, all baseline methods utilize the same stack of event frames $\mathcal{E}$ as input to estimate the impact timing.

\noindent\textbf{E2E-Spot~\cite{Hong_E2ESpot}:} A representative RGB-based action spotting method originally designed for standard video frames. Since this model is not optimized for event data, we fine-tune it on the generated event frames $\mathcal{E}$ using the same training protocol as our proposed method to adapt it to the event domain.

\noindent\textbf{SAM 3~\cite{SAM3}:} As there is no standard event-based method for this specific task, we utilize SAM 3, a state-of-the-art foundation model for image segmentation. We apply SAM 3 directly to the event frames $\mathcal{E}$ to generate segmentation masks. To ensure a fair comparison in spatial localization, we compute the weighted centroids from the output probability maps of SAM~3, following the same mechanism as our approach.

\noindent\textbf{YOLO 11~\cite{yolo11_ultralytics}:} We also employ YOLO 11, a state-of-the-art object detection framework. Similar to E2E-Spot, we fine-tune the model on the event frames $\mathcal{E}$. Since YOLO outputs bounding boxes, we select the detection with the highest confidence score and utilize the geometric center of this bounding box as the predicted spotting location.

\vspace{-2mm}
\begin{table}[t]
    \centering
    \caption{Quantitative evaluation results.}
    \label{tab:quantitative_evaluation}
    \resizebox{\linewidth}{!}{
    \begin{tabular}{l c c c}
        \toprule
        \textbf{Method} & \textbf{MAE} (ms) $\downarrow$ & \textbf{SR ($< 1\sigma$)} (\%) $\uparrow$ & \textbf{SR ($< 2\sigma$)} (\%) $\uparrow$ \\
        \midrule
        E2E-Spot & 17.2 & 10.0 & 15.0 \\
        SAM~3* & 3.09  & 30.0 & 52.5 \\
        YOLO~11* & 2.06  & 32.5 & 55.0 \\
        \midrule
        Ours     & \textbf{0.767} & \textbf{52.5} & \textbf{77.5} \\
        \bottomrule
        \multicolumn{4}{l}{\footnotesize * Methods adapted using geometric centers of masks/boxes.}
    \end{tabular}
    }
    \vspace{-4mm}
\end{table}

\begin{table}[t]
    \centering
    \caption{Ablation study on the proposed components.}
    \label{tab:ablation}
    \resizebox{\linewidth}{!}{
    \begin{tabular}{ccccc|c cc}
        \toprule
        \multicolumn{2}{c}{\textbf{Refine}} & & & & \multicolumn{2}{c}{\textbf{SR} (\%) $\uparrow$} \\
        \textbf{(fwd)} & \textbf{(bwd)} &
        \textbf{+$\mathcal{L}_{\text{circ}}$} & \textbf{+$\mathcal{L}_{\text{smooth}}$} &
        \textbf{MAE} (ms) $\downarrow$ & \textbf{$< 1\sigma$} & \textbf{$< 2\sigma$} \\
        \midrule
         & & & & 3.09 & 30.0 & 52.5 \\
        \checkmark &  &  &  & 2.665 & 37.5 & 62.5 \\
        \checkmark & \checkmark &  &  & 1.787 & 42.5 & 72.5 \\
        \checkmark & \checkmark & \checkmark &  & 1.696 & 47.5 & 72.5 \\
        \checkmark & \checkmark & \checkmark & \checkmark & \textbf{0.767} & \textbf{52.5} & \textbf{77.5} \\
        \bottomrule
    \end{tabular}%
    }
\end{table}

\vspace{-2mm}
\subsection{Evaluation Metrics}
\label{ssec:metrics}
\vspace{-2mm}

We evaluate the temporal precision using Mean Absolute Error (MAE) and Success Rate (SR).
For SR, we define the success thresholds based on human annotation variability to account for the inherent ambiguity of the impact timing.
Specifically, we calculate the standard deviation of the three annotators for each clip and average it across the dataset, resulting in a human baseline noise level of $\sigma = 0.513$ ms.
We report SR using tolerance thresholds of $1\sigma$ and $2\sigma$ ($1.026$ ms).

\vspace{-2mm}
\section{Results}
\label{sec:results}
\vspace{-2mm}

\subsection{Quantitative Evaluation}
\label{ssec:quantitative}
\vspace{-2mm}

As shown in Table~\ref{tab:quantitative_evaluation}, our method outperforms all baselines, achieving the best MAE of 0.767 and SR of 52.5\% ($<1\sigma$). These results correspond to an approximately 63\% reduction in MAE and a 20.0\% increase in SR compared to the second-best method, YOLO~11.
The ablation study in Table~\ref{tab:ablation} validates our design choices. The bidirectional refinement drastically drops the MAE from 2.665 to 1.787. Additionally, the geometric constraints ($\mathcal{L}_{\text{circ}}$ and $\mathcal{L}_{\text{smooth}}$) further refine the estimation, yielding the best performance across all metrics. (see supplementary material for per-scene MAE).

\begin{figure}[t]
  \centering
  \includegraphics[width=\linewidth]{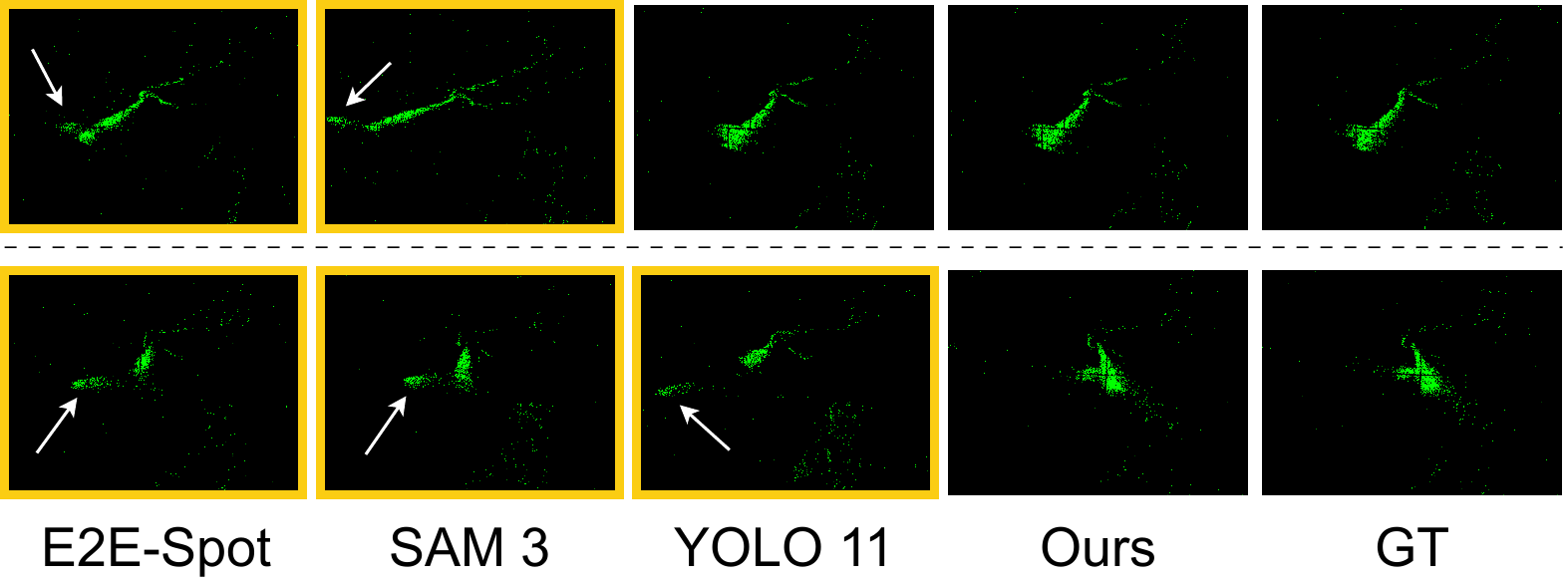}
  \vspace{-6mm}
  \caption{
  Qualitative comparison of estimated impact frames.
  Frames with \textbf{yellow borders} indicate failure cases where the error exceeds the $2\sigma$ threshold.}
  \label{fig:qualitative_result}
\end{figure}

\vspace{-2mm}
\subsection{Qualitative Evaluation}
\label{ssec:qualitative}
\vspace{-2mm}

Figure~\ref{fig:qualitative_result} visualizes the estimated impact frames to demonstrate the temporal precision.
Consistent with the success rate threshold defined in Sec.~\ref{ssec:metrics}, we mark predictions that deviate from the ground truth by more than $2\sigma$ with yellow borders.
The visualization highlights that while baseline methods frequently produce outliers indicated by yellow borders, our proposed method consistently estimates the impact timing within the tolerance range, aligning closely with the ground truth.

\vspace{-2mm}
\section{Conclusion}
\label{sec:conclusion}
\vspace{-2mm}

In this paper, we proposed the first high-temporal-resolution framework for estimating baseball impact timing using event-based cameras.
We constructed a novel dataset and introduced a pipeline that refines segmentation masks to calculate the weighted centroids of the ball and bat, enabling precise timing estimation.
Experimental results demonstrated that our method significantly outperforms baseline approaches.
Although our current approach assumes full visibility of the ball and bat and relies on a modestly sized dataset, we plan to address these limitations in future work by expanding the dataset and diversifying recording conditions.
\vspace{1mm}

\noindent
\textbf{{Acknowledgement.}}
\noindent
This work was partially supported by JST Presto JPMJPR22C1, JSPS Grant-in-Aid for Challenging Research (Exploratory) 24K22296.
Wataru Ikeda was supported by JST BOOST, Japan Grant Number JPMJBS2409.
\clearpage
\bibliographystyle{IEEEbib}
\bibliography{refs}
\end{document}


%
\maketitle

\hypersetup{linkcolor=black}
\tableofcontents
\hypersetup{linkcolor=red}

\section{Overview of the Supplementary Material}
This supplementary material provides additional details that could not be included in the main paper due to space constraints. 
In Section~\ref{sec:supp_experiment}, we visualize the experimental setup used for data collection and validate our ground truth strategy.
Section~\ref{sec:supp_ablation} presents a detailed quantitative evaluation, including per-scene performance breakdowns and ablation studies.
Section~\ref{sec:implementation_details} describes the implementation details, including hyperparameters and hardware specifications. 
Finally, Section~\ref{sec:supp_related_work} provides an extended discussion on related work, a visual validation of baseline limitations, and a failure analysis of our method. 

\section{Experimental Setup and GT Validation}
\label{sec:supp_experiment}

\subsection{Physical Configuration}
\begin{figure}[t]
  \centering
  \includegraphics[width=\linewidth]{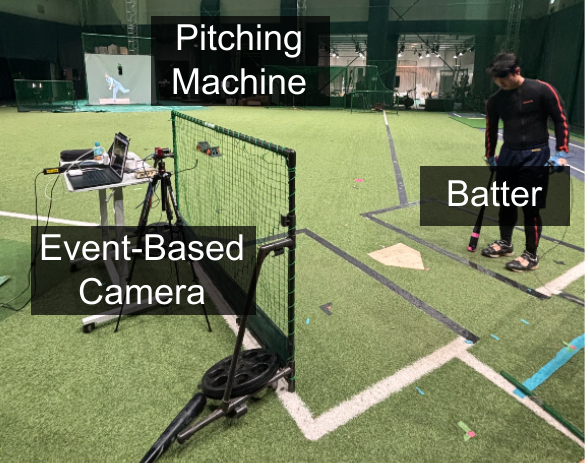} 
  \vspace{-2mm}
  \caption{Experimental setup used for data collection. A pitching machine delivers balls to a batter, while an event camera records the side view to capture high-speed motion.}
  \label{experiment_image}
\end{figure}

The physical configuration of our data collection environment is illustrated in \Cref{experiment_image}. 
We utilized a standard pitching machine to deliver balls to a batter within an indoor facility. 
An event camera was positioned to capture the side view of the trajectory to record high-temporal-resolution data.

\subsection{Validation of Ground Truth Strategy}
To justify the necessity of manual annotation over sensor-based automation, we conducted a preliminary validation using an IMU attached to the bat knob. The IMU signal was sampled at 1000 Hz, while the event camera recorded at 10,000 fps. The sensors were hardware-synchronized via an LED trigger.

\noindent\textbf{IMU Impact Detection.} 
We defined the impact timing from the IMU signal based on the magnitude of the acceleration vector. Specifically, the impact frame $t_{\text{IMU}}$ was identified as the moment where the squared norm of the acceleration $(x^2+y^2+z^2)$ exceeded twice the value of the preceding frame. This thresholding method was chosen to detect the sharp initial shock of the impact.

\begin{figure}[h]
  \centering
  \includegraphics[width=\linewidth]{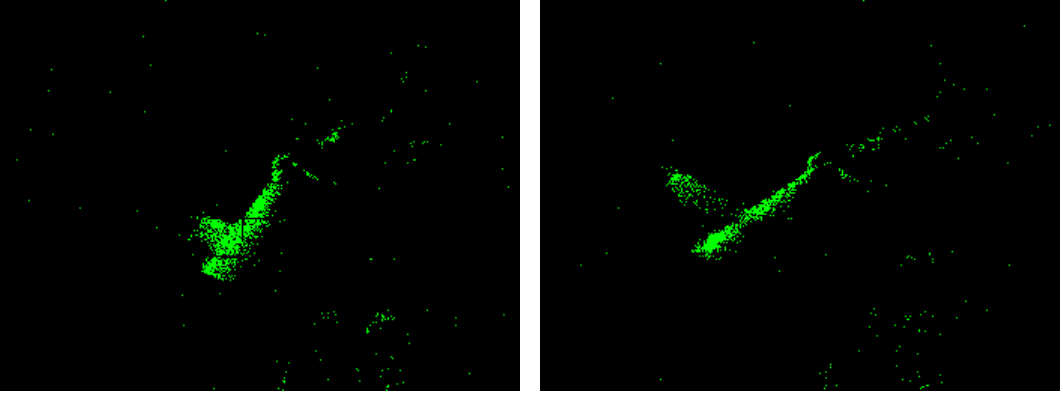} 
  \vspace{-2mm}
  \caption{Visual comparison at the maximum IMU latency (+7.07 ms). \textbf{Left:} The true impact moment annotated manually ($t_{\text{GT}}$). \textbf{Right:} The frame when the IMU triggered ($t_{\text{IMU}}$). The ball has already moved significantly away from the bat, demonstrating that the IMU signal is too slow for sub-millisecond evaluation.}
  \label{fig:latency_analysis}
\end{figure}

\noindent\textbf{Latency Analysis.}
We compared $t_{\text{IMU}}$ against the manual ground truth $t_{\text{GT}}$ obtained from the high-speed event stream. Across 28 valid trials, the time lag ($t_{\text{IMU}} - t_{\text{GT}}$) exhibited a significant, variable latency ranging from 0.40 ms to 7.07 ms (Mean: $3.32 \pm 2.12$ ms). This stochastic delay is likely due to mechanical propagation within the bat and internal sensor buffering. Since our method requires sub-millisecond accuracy, this jitter renders the IMU signal insufficient as a ground truth (\Cref{fig:latency_analysis}). Consequently, we adopted manual annotation on 10,000 fps frames to ensure the highest temporal precision.

\section{Additional Quantitative Results}
\label{sec:supp_ablation}

\subsection{Per-Scene Evaluation}
While the main paper reports the average performance across all datasets, Table~\ref{tab:evaluation_detailed} presents the detailed Mean Absolute Error (MAE) for each of the seven scenes individually. 
The results indicate that our method achieves consistent performance across varying ball speeds and trajectories. In particular, while baseline methods such as E2E-Spot and SAM~3 struggle significantly in complex scenarios (e.g., Scene 4 and Scene 5), our approach maintains a low error rate, demonstrating its robustness.

\begin{table*}[t]
    \centering
    \caption{Detailed quantitative evaluation per scene (MAE in ms). Our method outperforms baselines in most scenarios and maintains stability in challenging scenes (e.g., scene 5).}
    \label{tab:evaluation_detailed} 
    \resizebox{\linewidth}{!}{%
    \begin{tabular}{lccccccc|c} 
        \toprule
         & \textbf{scene1} & \textbf{scene2} & \textbf{scene3} & \textbf{scene4} & \textbf{scene5} & \textbf{scene6} & \textbf{scene7} & \textbf{Avg.}\\
        \midrule
        E2E-Spot & 6.907 & 10.708 & 2.811  & 21.856 & 25.511 & 30.419 & 18.413 & 17.204\\
        SAM~3 & 0.594 & 5.568 & 1.915 & 2.03 & 7.95 & 1.706 & 2.448 & 3.087\\
        YOLO~11 & 0.574 & \textbf{0.333} & 2.915 & 0.946  & 6.1 & 1.866 & \textbf{1.206} & 2.056\\
        \hline
        Ours & \textbf{0.366} & 0.433 & \textbf{0.338} & \textbf{0.841} & \textbf{1.117} & \textbf{0.666} & 1.5 &\textbf{0.767}\\
        \bottomrule
    \end{tabular}%
    }
\end{table*}

\subsection{Ablation Studies}

\noindent\textbf{Component Analysis.}
Table~\ref{tab:ablation_detailed} analyzes the contribution of each component in our proposed pipeline. 
The baseline with only forward refinement shows limited accuracy. Incorporating bidirectional refinement (Refine bwd) significantly reduces the MAE, proving the effectiveness of leveraging temporal context from both directions. 
Furthermore, the addition of the geometric circle loss ($\mathcal{L}_{\text{circ}}$) and smoothness loss ($\mathcal{L}_{\text{smooth}}$) further refines the trajectory, leading to the best performance.

\begin{table*}[t]
    \centering
    \caption{Ablation study on method components. ``Refine (fwd/bwd)" denotes the refinement modules, and $\mathcal{L}_{\text{circ/smooth}}$ represents the auxiliary loss terms.}
    \label{tab:ablation_detailed} 
    \resizebox{\linewidth}{!}{%
    \begin{tabular}{cccc|ccccccc|c} 
        \toprule
        \textbf{+ Refine (fwd)} & \textbf{+ Refine (bwd)} &
        \textbf{+ $\mathcal{L}_{\text{circ}}$} & \textbf{+ $\mathcal{L}_{\text{smooth}}$}& \textbf{scene1} & \textbf{scene2} & \textbf{scene3} & \textbf{scene4} & \textbf{scene5} & \textbf{scene6} & \textbf{scene7} &
        \textbf{Avg.}\\
        \midrule
         &  &  &  & 0.594 & 5.568 & 1.915 & 2.03 & 7.95 & 1.706 & \underline{2.448} & 3.087 \\
        \checkmark &  &  &  & 0.414 & 4.768 & 2.962 & 4.212 & 1.15 & 1.834 & 2.74 & 2.665 \\
        \checkmark & \checkmark &  &  & \textbf{0.354} & \underline{1.608} & 0.522 & 0.919 & 1.16 & \textbf{0.57} & 8.412 & 1.787 \\
        \checkmark & \checkmark & \checkmark &  & 0.474 & 1.617 & \textbf{0.198} & \textbf{0.801} & \textbf{1.06} & 0.966 & 7.88 & \underline{1.696} \\
        \checkmark & \checkmark & \checkmark &  \checkmark & \underline{0.366} & \textbf{0.433} & \underline{0.338} & \underline{0.841} & \underline{1.117} & \underline{0.666} & \textbf{1.5} & \textbf{0.767} \\
        \bottomrule
    \end{tabular}%
    }
\end{table*}

\vspace{2mm}
\noindent\textbf{Impact of Class Balancing Weights.}
Given the small size of the ball target, appropriate weighting for the Cross-Entropy loss is crucial. 
Table~\ref{tab:ablation_study} investigates the impact of varying the ball class weight $\lambda_{\text{ball}}$. 
Standard weighting ($\lambda_{\text{ball}}=1.0$) results in high MAE due to the dominance of the background class. Increasing $\lambda_{\text{ball}}$ improves the Success Rate (SR). We empirically found that $\lambda_{\text{ball}}=13.0$ yields the optimal trade-off, minimizing MAE while maintaining a consistently high SR. Excessive weighting (e.g., 20.0) degrades performance, likely due to an increase in false positives.

\begin{table}[h]
    \centering
    \caption{Ablation study on the loss weights for background, bat, and ball classes. We investigate the impact of varying $\lambda_{\text{ball}}$ while fixing the others. The most balanced performance is archived when $\lambda_{\text{ball}} = 13$.}
    \label{tab:ablation_study}
    \resizebox{\linewidth}{!}{ 
    \begin{tabular}{c c c | c c c}
        \toprule
        \multicolumn{3}{c|}{\textbf{Weights}} & \multicolumn{3}{c}{\textbf{Metrics}} \\
        $\lambda_{\text{bg}}$ & $\lambda_{\text{bat}}$ & $\lambda_{\text{ball}}$ & \textbf{MAE} (ms) $\downarrow$ & \textbf{SR ($< 1\sigma$)} (\%) $\uparrow$ & \textbf{SR ($< 2\sigma$)} (\%) $\uparrow$ \\
        \midrule
        1.0 & 1.0 & 1.0 & 2.523 & 40.0 & 75.0 \\ 
        0.5 & 1.0 & 1.0 & 2.139 & 42.5 & 72.5 \\  
        0.5 & 1.0 & 10.0 & 1.54 & \textbf{57.5} & \textbf{77.5} \\
        0.5 & 1.0 & 13.0 & \textbf{0.767} & 52.5 & \textbf{77.5} \\ 
        0.5 & 1.0 & 20.0 & 2.192 & 47.5 & 72.5 \\ 
        \bottomrule
    \end{tabular}
    }
\end{table}

\vspace{2mm}
\noindent\textbf{Parameter Sensitivity.}
We also evaluated the network's sensitivity to the event accumulation parameter. When testing different accumulation values of 5, 10, and 20, the resulting MAEs were 2.93, 0.77, and 1.65, respectively. This confirms that our chosen value of 10 is optimal for capturing the necessary temporal information while maintaining precision.

\section{Implementation Details}
\label{sec:implementation_details}

Our method was implemented using PyTorch. The input frames were resized to a resolution of $256 \times 320$ pixels.
The network was trained using the Adam~\cite{Kingma_Adam} optimizer with an initial learning rate of $5 \times 10^{-4}$, a batch size of 8, and a total of 50 epochs.
The balancing weights for the total loss function were set to $\lambda_{\text{dice}}=1.0$, $\lambda_{\text{ce}}=0.5$, $\lambda_{\text{smooth}}=0.1$, and $\lambda_{\text{circle}}=0.05$. 

To address the severe class imbalance inherent in the dataset where the ball occupies a very small portion of the frame compared to the background, we applied weighted Cross-Entropy loss. The class weights were assigned as 0.5, 1.0, and 13.0 for the background, bat, and ball classes, respectively. The selection process for these weights is discussed in Section~\ref{sec:supp_ablation}.
All experiments were conducted on a workstation equipped with an NVIDIA GeForce RTX 4070 GPU. Under this environment, the inference time for our proposed module is exceptionally fast at just 3 ms per frame. However, the overall speed depends on the preprocessing step required to generate the raw mask via the segmentation module. For instance, using SAM3 takes approximately 150 ms per frame.
Regarding the data preparation, the manual annotation burden is relatively light. It takes approximately 30 sec to create a ground truth label for a single frame.

\begin{figure*}[t]
  \centering
  \includegraphics[width=0.9\linewidth]{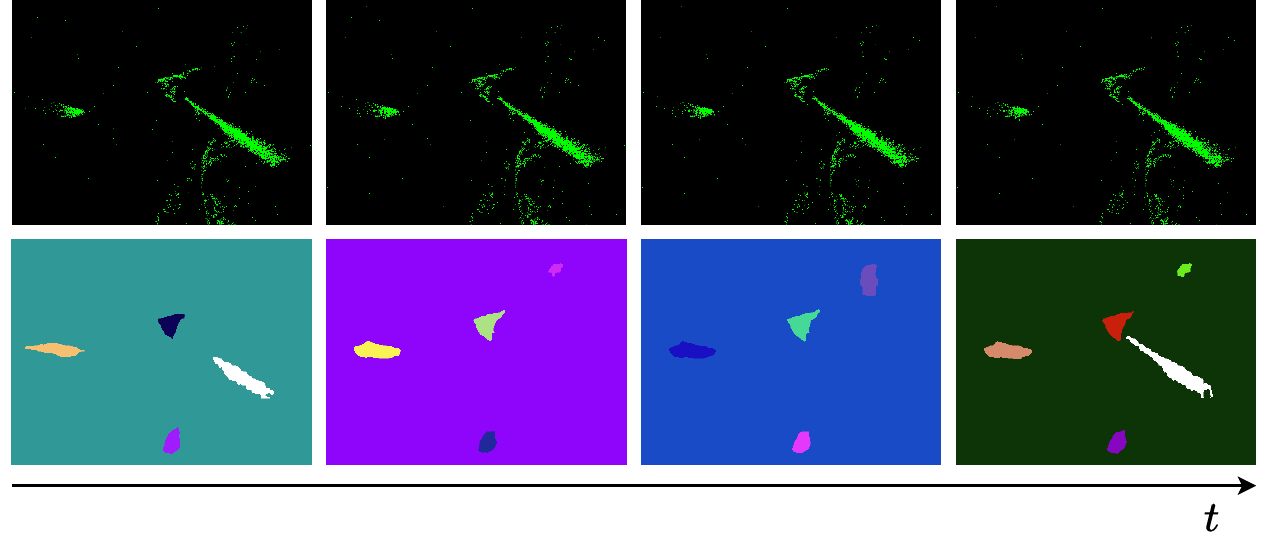} 
  \vspace{-2mm}
  \caption{Visual verification of the limitations of EventSAM~\cite{Chen_EventSAM}.
  \textbf{Top Row:} Input densified event frames. \textbf{Bottom Row:} Corresponding segmentation masks from EventSAM.
  The baseline exhibits temporal instability and susceptibility to environmental noise, confirming the necessity of our proposed bidirectional framework.}
  \label{fig:eventsam_failure}
\end{figure*}
\section{Extended Related Work and Discussion}
\label{sec:supp_related_work}

This section provides a detailed comparison with existing literature, focusing on technical distinctions and domain-specific applications.

\subsection{Comparison with Existing Event-based Methods}
Regarding the novelty of our event representation and bidirectional reasoning, we provide further clarification against recently suggested works:
\begin{itemize}
    \item \textbf{Event Representation:} Unlike motion-compensated tensors~\cite{Yao_Rethinking} or static 6-channel statistics~\cite{Alonso_EVSegNet}, our dense event frames use sliding window accumulation. This overlapping approach maintains the high temporal resolution critical for capturing high-speed dynamics.
    \item \textbf{Bidirectional Reasoning:} While some automotive segmentation models~\cite{Yao_Rethinking, Annamalai_EventMASK} leverage temporal context, they primarily focus on feature propagation for mask consistency. Our novelty lies in using bidirectional tracking specifically for precise high-speed object time estimation.
    \item \textbf{Asynchronous Learning and Tracking:} Other approaches like EvDistill~\cite{Wang_Evdistill} use cross-modal distillation, and TETO~\cite{Yang_TETO} focuses on motion estimation. While relevant, our framework prioritizes a lightweight refinement-based approach for impact detection.
\end{itemize}

\subsection{Domain-Specific Applications and Hardware Considerations}
Several studies have explored event cameras in sports, particularly for ball spin estimation~\cite{Nakabayashi_Spin, Gossard_TableTennis}. While these works share the domain, our focus remains on the precise timing of impact events rather than rotational dynamics. 
Furthermore, while global shutter cameras offer advantages in certain scenarios, the inherent limitations compared to event cameras—such as motion blur and lower temporal resolution—justify our event-based approach for high-speed analysis, as discussed in the experimental comparisons by Holešovský et al.~\cite{Holesovsky_Sensors}.

\subsection{Visual Validation of Baseline Limitations}
\label{ssec:supp_eventsam}
As a qualitative supplement to the quantitative analysis, we further examine the limitations of existing adaptations like EventSAM~\cite{Chen_EventSAM}. \Cref{fig:eventsam_failure} provides a visual breakdown of its failure modes in high-speed scenarios.

As evident in the figure, since EventSAM processes each frame independently, the predicted masks exhibit significant temporal jitter. The bat mask fluctuates violently in shape or disappears entirely between consecutive frames, failing to track the object reliably. This lack of temporal context exacerbates susceptibility to environmental noise, confirming that frame-wise segmentation without domain-specific temporal adaptation is insufficient for robust microsecond-level analysis.

\subsection{Failure Analysis of the Proposed Method}
\label{sec:failure_analysis}
\begin{figure}[t]
  \centering
  \includegraphics[width=\linewidth]{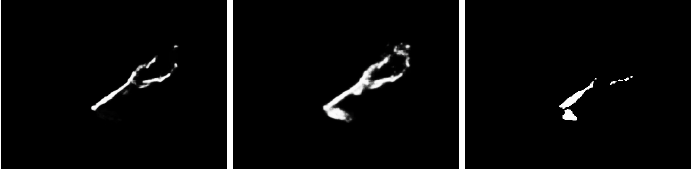}
  \caption{A typical failure case of our refinement module. \textbf{Left / Center:} Raw bat masks from forward and backward tracking (using SAM 3) exhibiting severe noise. \textbf{Right:} The resulting refined mask. When significant artifacts exist simultaneously in both directional priors, the residual noise propagates to the final mask, decreasing the accuracy of the impact timing estimation.}
  \label{fig:failure_analysis}
\end{figure}

While our bidirectional refinement framework is generally robust against environmental noise, it can still fail when the initial segmentation masks are heavily degraded. \Cref{fig:failure_analysis} illustrates a typical failure case. If the raw masks generated by the baseline contain severe artifacts in both the forward and backward tracking directions, our refinement module may not fully eliminate this simultaneous noise. Consequently, the residual noise in the finalized mask can distort the object's spatial representation, leading to a shifted estimation of the impact timing.

\bibliographystyle{IEEEbib}
\bibliography{refs}